\pdfoutput=1

\documentclass[11pt]{article}

\usepackage[preprint]{acl}

\usepackage{times}
\usepackage{latexsym}

\usepackage[T1]{fontenc}

\usepackage[utf8]{inputenc}

\usepackage{microtype}

\usepackage{inconsolata}

\usepackage{graphicx}

\usepackage{comment}
\usepackage{amssymb}
\usepackage{graphicx}
\usepackage{multirow}
\usepackage{amsmath}

\title{ROSAQ: Rotation-based Saliency-Aware Weight Quantization for Efficiently Compressing Large Language Models}

\author{
Junho Yoon \\ Jeonbuk National University \\ \texttt{hoho5702@jbnu.ac.kr} \\ \And
Geom Lee \\ Jeonbuk National University \\ \texttt{lglglglg@jbnu.ac.kr} \\ \And
Donghyeon Jeon \\ Naver \\ \texttt{donghyeon.jeon@navercorp.com} \\ \AND
Inho Kang \\ Naver \\ \texttt{once.ihkang@navercorp.com} \\ \And
Seung-Hoon Na \\ UNIST \\ \texttt{nash@unist.ac.kr} \\
}

\begin{document}
\maketitle
\begin{abstract}
\textit{Quantization} has been widely studied as an effective technique for reducing the memory requirement of large language models (LLMs), potentially improving the latency time as well. Utilizing the characteristic of \textit{rotational invariance} of transformer, we propose the \underline{ro}tation-based \underline{sa}liency-aware weight \underline{q}uantization (\textbf{ROSAQ}), which identifies salient channels in the projection feature space, not in the original feature space, where the projected ``principal’’ dimensions are naturally considered as ``salient’’ features. The proposed ROSAQ consists of 1) \textit{PCA-based projection}, which first performs principal component analysis (PCA) on a calibration set and transforms via the PCA projection, 2) \textit{Salient channel identification}, which selects dimensions corresponding to the $K$-largest eigenvalues as salient channels, and 3) \textit{Saliency-aware quantization} with mixed-precision, which uses FP16 for salient dimensions and INT3/4 for other dimensions. Experiment results show that ROSAQ shows improvements over the baseline saliency-aware quantization on the original feature space and other existing quantization methods. With kernel fusion, ROSAQ presents about 2.3x speed up over FP16 implementation in generating 256 tokens with a batch size of 64.
\end{abstract}

\section{Introduction}
Large language models (LLMs) have shown remarkable performances on various NLP tasks, however, due to their huge sizes of model parameters, LLMs require significant GPU memory for inference, substantially limiting the throughput and latency time. 
To address these challenges, \textit{quantization} methods \cite{zeroquant, llm.int8, qlora, zeroquant-fp, zeroquant-v2, squeezellm} have been widely studied as an effective technique for reducing the memory requirement of LLMs, potentially improving the latency time as well, by representing weights and activations of LLMs using low-precision. 

In quantization, one of the most challenging issues is the presence of \textit{outliers} in weights and activations, as they widen the quantization range and increase quantization error. Recently, leveraging the \textit{rotational invariance} of transformers \cite{spinquant}, rotation-based quantization has been extensively applied to mitigate outliers, motivated by the observation that outliers are reduced after rotation. Similarly, SmoothQuant \cite{smoothquant} exploits the \textit{scaling invariance} of linear layer, by dividing activation values by channel-specific scaling factors, thus greatly reducing activation outliers without severely strengthening weight outliers. The scaling invariance is further utilized in activation-aware quantization (AWQ) \cite{awq}, which primarily focuses on ``salient channels’’ to reduce quantization errors, by identifying salient channels based on activation magnitude.

Without being limited in the original feature space as in \cite{awq}, this paper extensively explores the rotational invariance for \textit{saliency-aware weight quantization} by identifying salient channels based on ``principal dimensions on the projection space,” thereby proposing the \underline{ro}tation-based \underline{sa}liency-aware weight \underline{q}uantization (\textbf{ROSAQ}). By the definition, the principal dimensions resulting from the principal component analysis (PCA) maximize the variances of channel values on projected space, and accordingly substantially increase their activation magnitudes. Our key underlying expectation is that these principal channels with the largest eigenvalues are more dominant and salient than the existing magnitude-based salient channels in original space, due to their inherent properties of maximizing variance, thereby further improving the saliency-aware quantization. The proposed ROSAQ consists of three steps:
\begin{itemize}
\item \textbf{PCA-based projection}, which first performs PCA projection with its eigenvectors on a calibration set to obtain the PCA-projected calibration set. For the multi-head self-attention (MHSA) layer, we further propose the use of \textit{head-wise} PCA, where the PCA projection is applied  separately to each head-specific attention representation.
\item \textbf{Salient channel identification}, which selects ``principal channels’’ corresponding to the $K$-largest eigenvalues as `salient’ channels, and regards the other channels as normal non-salient channels.
\item \textbf{Saliency-aware quantization with mixed-precision}, which applies the \textit{per-group} quantization, where employs FP16 for a salient group of channels, and INT3/4 for all other groups of non-salient channels, where a group consists of 128 channels.
\end{itemize}

Experiment results on Wikitext2, zero-shot common-sense reasoning, and zero-shot MMLU tasks show that the proposed ROSAQ leads to improvements over the baseline saliency-aware quantization with mixed-precision on original feature space and the existing quantization methods, with minimal performance degradation. Furthermore, with kernel fusion, ROSAQ exhibits about 2.3x speed up over FP16 implementation when generating 256 tokens with a batch size of 64, and about 2x speedup when generating 128 tokens with a batch size of 128.

Our contributions are summarized as follows: 1) we propose ROSAQ, which is a novel rotation-based saliency-aware quantization, by choosing principal channels resulting from the PCA projection as salient ones, 2) we apply the head-wise PCA projection across multiple heads for quantizing the parameters of MHSA, and 3) the proposed ROSAQ leads to improved performances over existing quantization on Wikitext2, zero-shot common-sense reasoning, and zero-shot MMLU tasks. 

\section{Related Work} 
Quantization methods have been studied mainly two categories -- quantization-aware training (QAT) \cite{liu2023llmqatdatafreequantizationaware, shen2024edgeqatentropydistributionguided, ma2024era1bitllmslarge} and post-training quantization (PTQ) \cite{smoothquant+, quipsharp, zeroquant42, aptq, exploreptq, qllm, spqr, omniquant}. PTQ is widely applied because it requires no (or minimal) training and only needs to use a small calibration set. While the ``rotational invariance'' has previously applied for the language model pruning \cite{slicegpt, sp3}, the rotation-based quantization has been extensively studied to reduce outliers, including incoherence processing based on orthogonal projections \cite{quip, quarot}, and optimizing the rotation matrix \cite{spinquant} based on Cayley SGD\cite{cayleysgd}. 

Saliency-aware quantization has also been proposed by AWQ \cite{awq}, which selects salient channels based on activation magnitudes but uses ``full low-precision'' quantization, leveraging the ``scaling invariance'' property of linear layers without mixed precision. 

Unlike rotation-based quantization methods such as \cite{spinquant}, which aim to remove outliers, ROSAQ applies rotation to more effectively identify salient channels. Instead of the scaling invariance used in AWQ, ROSAQ exploits rotational invariance for salient channel identification.

\begin{figure}[t!]
\centering
\includegraphics[width=7cm]{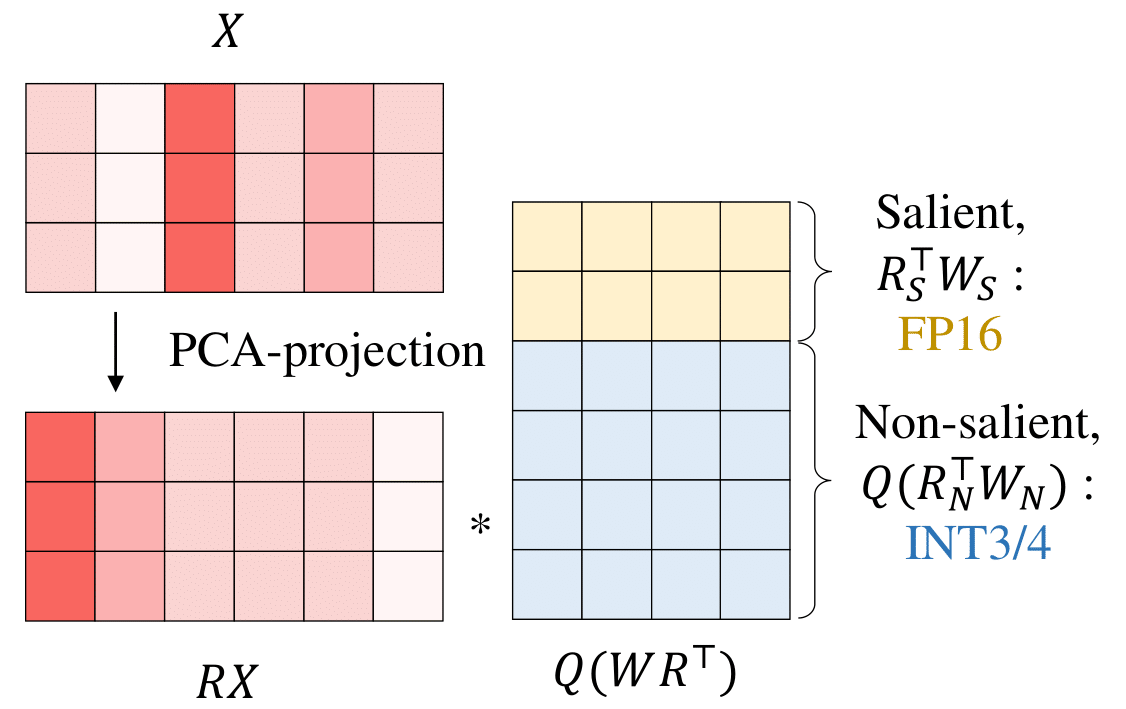}
\caption{An overview diagram of ROSAQ that quantizes the weights of a linear layer ${\mathbf{X}} {\mathbf{W}}$, using rotational invariance as described by Eq. (\ref{eq_rotational_invarience_linear_layer}), where ${\mathbf{X}}$ is the calibration data matrix. ROSAQ first applies the PCA-based projection, taking ${\mathbf{Q}}$ as ${\mathbf{R}}$, with eigenvectors obtained from Eq. (\ref{eq_PCA_calibration_set}). The salient channels denoted as ${\mathbf{W}}_S$, corresponding to the $K$ largest eigenvalues, are represented in FP16, while the remaining non-salient channels ${\mathbf{W}}_N$ are represented in low precision, such as INT3/INT4.
}
\label{fig:overview}
\end{figure}

\section{Method: ROSAQ}
Figure \ref{fig:overview} presents the overall diagram of ROSAQ. Suppose that a calibration set consists of $N$ data samples, formally presented as ${\mathbf{X}}=\left[ {\mathbf{x}}_1, \cdots, {\mathbf{x}}_N \right]^T \in {\mathbb{R}}^{N \times d}$, where ${\mathbf{x}}_i \in {\mathbb{R}}^{d}$ is $i$-th data representation. ROSAQ uses the rotational invariance for all linear layers formed with ${\mathbf{X}}{\mathbf{W}}$ in transformer, where ${\mathbf{W}} \in {\mathbb{R}}^{d \times d’} $ is a weight matrix of parameters, formulated as follows:
\begin{eqnarray}
{\mathbf{X}}{\mathbf{W}} = \left( {\mathbf{X}}{\mathbf{R}}\right) \left( {\mathbf{R}}^T{\mathbf{W}} \right) 
\label{eq_rotational_invarience_linear_layer}
\end{eqnarray}
where ${\mathbf{R}} \in {\mathbb{R}}^{d \times d}$ is a \textit{rotation} matrix, which consists of orthonormal vectors. 

After applying the weight quantization, Eq. (\ref{eq_rotational_invarience_linear_layer}) is approximated by:
\begin{eqnarray}
{\mathbf{X}}{\mathbf{W}} \approx \left( {\mathbf{X}}{\mathbf{R}}\right) Q\left( {\mathbf{R}}^T{\mathbf{W}} \right) 
\label{eq_weight_quantization_linear_layer}
\end{eqnarray}
where $Q$ is a quantization function, which adopts \textit{per-group} quantization with a group size of 128 channels.

Similar to AWQ \cite{awq}, ROSAQ also takes into account the assumption that weights are not equally salient. Different from AWQ which applies the quantization to all channels, but in a scale-sensitive manner, ROSAQ deploys the mixed-precision which keeps high-precision for salient channels while using low-precision non-salient channels, in order to minimize the quantization error particularly for salient channels. To formally present the mixed-precision, suppose the column vectors of ${\mathbf{R}}$ are sorted by their saliency degrees, and they are then divided into two groups – salient and non-salient groups of channels, ${\mathbf{R}}=\left[ {\mathbf{R}}_S, {\mathbf{R}}_N \right]$, where ${\mathbf{R}}_S \in {\mathbb{R}}^{K}$ is the orthonormal vectors for salient channels, and ${\mathbf{R}}_N \in {\mathbb{R}}^{d -K}$ is one for non-salient channels.
Under the mixed-precision, Eq. \ref{eq_weight_quantization_linear_layer} is approximated by:
\begin{eqnarray}
{\mathbf{X}}{\mathbf{W}} \approx  \left( {\mathbf{X}}{\mathbf{R}}\right) \left[ \begin{array}{c}\left({\mathbf{R}}_S^T{\mathbf{W}}_S \right) \\
Q\left( {\mathbf{R}}_N^T{\mathbf{W}}_N \right)
\end{array}
\right] 
\label{eq_weight_quantization_mixed_precision_linear_layer}
\end{eqnarray}
where ${\mathbf{W}}_S \in {\mathbb{R}}^{K \times d'}$ is the sub-block of the weight matrix for salient channels, and  ${\mathbf{W}}_N \in {\mathbb{R}}^{(d-K) \times d'}$ is one for non-salient channels. 

ROSAQ consists of three steps –- 1) PCA-based projection, 2) Salient channel identification, and 3) Saliency-aware quantization with mixed-precision, which will be presented in the next subsections with more details.

\begin{figure}[!htbp]
\centering
\includegraphics[width=7cm]{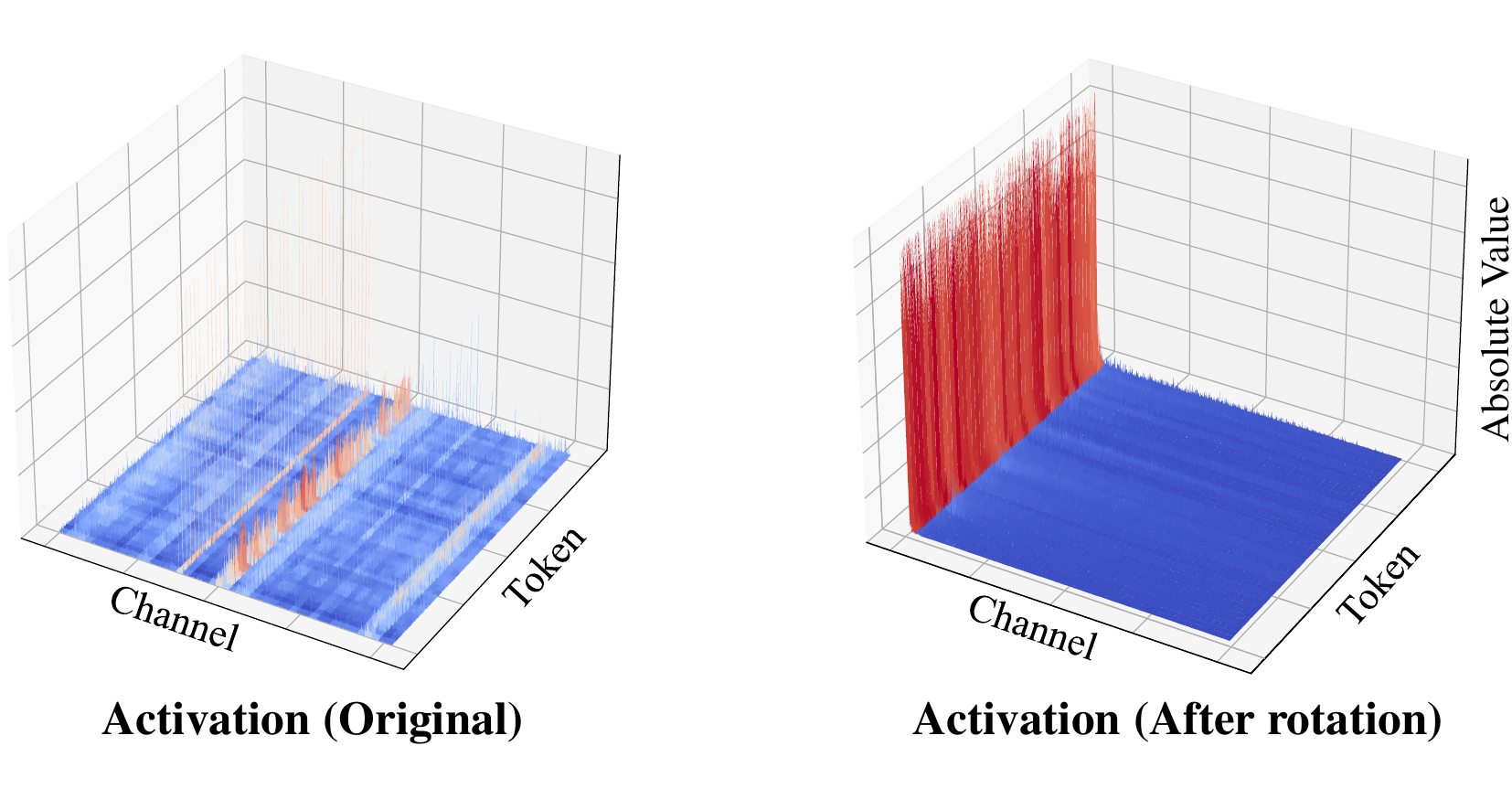}
\caption{
Magnitude of the input activation values to MHSA in LLaMA2-7B, before and after PCA-based rotation. Salient channels are more dominant in PCA-projected space than the original space. Detailed statistics are presented in Appendix \ref{Appendix:Distribution}.
}
\label{fig:activation_change}
\end{figure}

\subsection{PCA-based projection for Computing ${\mathbf{R}}$}\label{sec:PCA-based}

To obtain the rotation matrix $\mathbf{R}$ in Eq. \ref{eq_weight_quantization_mixed_precision_linear_layer}, we perform PCA on the calibration set ${\mathbf{X}}$ as follows:
\begin{eqnarray}
{\mathbf{X}}^T {\mathbf{X}} = {\mathbf{R}}{\mathbf{\Lambda}} {\mathbf{R}}^T
\label{eq_PCA_calibration_set}
\end{eqnarray}
where ${\mathbf{R}} \in {\mathbb{R}}^{d \times d}$ is the eigenvectors of ${\mathbf{X}}^T {\mathbf{X}}$, and ${\mathbf{\Lambda}}$ is the corresponding eigenvalue matrix. Without loss of generality, we assume that the column vectors of ${\mathbf{R}}$ are sorted by their eigenvalues. 

To check whether the PCA projection helps to identify salient channels, Fig \ref{fig:overview} presents the activation magnitudes across all channels in both the original and PCA-projected feature spaces. 
It is  seen that the activation values for salient channels are more dominant, making them easier to distinguish compared to those in the original feature space.

\subsubsection{Head-wise PCA Projection in MHSA}\label{sec:Head-wise}
In general, ROSAQ uses the layer-wise PCA projection, which is applied individually to the activation matrix ${\mathbf{X}}_l$ in a layer-specific manner, for each linear layer represented by ${\mathbf{X}}_l {\mathbf{W}}_l$, thereby resulting in its own rotation matrix ${\mathbf{R}}_l$ for each layer $l$. 

To better capture the head-specific characteristics for quantization, ROSAQ deploys a head-wise PCA projection for MHSA, where a separate PCA is performed for each head-specific attentive representation. More specifically, suppose that ${\mathbf{H}}_h \in {\mathbb{R}}^{m \times d_h}$ is $h$-th head-specific attentive representation resulting from the activation matrix ${\mathbf{Z}}_{l-1}$ at the previous layer, as follows. 
\begin{eqnarray}
{\mathbf{H}}_h = \mbox{Attention}\left( {\mathbf{Z}}_{l-1}{\mathbf{W}}_h^Q,  {\mathbf{Z}}_{l-1}{\mathbf{W}}_h^K, {\mathbf{Z}}_{l-1}{\mathbf{W}}_h^V    \right) 
\label{eq_head_specific_self_attention}
\end{eqnarray} 
where ${\mathbf{W}}_h^Q, {\mathbf{W}}_h^K, {\mathbf{W}}_h^V \in {\mathbb{R}}^{d \times d_h}$ are the weight matrices at $h$-th head for query, key, and value parts, respectively. 

Instead of applying a global PCA on the concatenated multi-head representation $\mbox{concat}\left( {\mathbf{H}}_1, \cdots, {\mathbf{H}}_H \right)$, we approximate MHSA by using the head-specific PCA projection as follows:
\begin{eqnarray}
\mbox{MHSA}\left({\mathbf{Z}}_{l-1} \right) \approx \sum_{h=1}^{H} \left( {\mathbf{H}}_h {\mathbf{R}}_h \right) Q\left( {\mathbf{R}}_h^T {\mathbf{W}}_{h}^O  \right)
\label{eq_head_specific_PCA}
\end{eqnarray} 
where ${\mathbf{R}}_h$ is a head-specific PCA projection matrix, which consists of eigenvectors obtained by applying PCA on the head-specific calibration set ${\mathbf{X}}_{h} \in {\mathbb{R}}^{N \times d_h}$ for $h$-th head, and ${\mathbf{W}}_{h}^O \in {\mathbb{R}}^{d_h \times d}$ is the output projection matrix. In Appendix \ref{Appendix:Headwise rotation}, we present that the use of head-specific PCA leads to the decrease of perplexity, comparing to the global PCA, as in Table \ref{tab:head-wise}. 

\subsection{Salient channel identification}
\label{section_salient_channel_identification}
To identify salient channels on the PCA-based projection space, we sort projected channels according to their eigenvalues, and select a group of channels corresponding to the larger eigenvalues as salient ones. In Appedix \ref{Appendix:Distribution}, Table \ref{tab:magnitude} shows that the top-ranked channels also tend to have the largest average magnitudes. 

\subsection{Saliency-aware quantization with mixed-precision channel}
After splitting weights into two groups -- ${\mathbf{W}}_S$ and ${\mathbf{W}}_N$ -- salient and non-salient channels, we retain FP16 precision for the salient group, while applying quantization under the INT3/INT4 to the non-salient group. Detailed settings can be found in Appendix \ref{Appendix:weightsplitting}.

\begin{table}[t!]
\centering
\resizebox{0.48\textwidth}{!}{
\begin{tabular}{ccrrr}
\hline
\textbf{Model}  & \textbf{Method} & \textbf{PPL} ($\downarrow$) & \textbf{CSR Avg. ($\uparrow$)} & \textbf{MMLU} ($\uparrow$) \\ 
\hline
\multirow{6}{*}{\centering Llama3-8B} & FP16      & 6.14          & 68.31           & 62.10          \\ \cline{2-5}
                                      & GPTQ      & *6.50         & 67.03           & 59.86          \\
                                      & SpinQuant & 6.71          & 66.20           & 59.00          \\
                                      & AWQ       & 6.53          & 67.34           & 60.48          \\
                                      & Mixed     & 6.63          & 67.28           & 59.80          \\
                                      & ROSAQ     & \textbf{6.50} & \textbf{67.72}  & \textbf{60.97} \\ 
\hline
\multirow{5}{*}{\centering Qwen2-7B}  & FP16      & 7.14          & 68.92           & 69.42          \\ \cline{2-5}
                                      & GPTQ      & 7.30          & 68.76           & 67.26          \\
                                      & AWQ       & 7.33          & 68.74           & 68.79          \\
                                      & Mixed     & 7.39          & 68.29           & 67.81          \\
                                      & ROSAQ     & \textbf{7.29} & \textbf{68.87}  & \textbf{68.82} \\ 
\hline
\end{tabular}
}
\caption{
Comparison results between ROSAQ and other quantization methods on LLaMA3-8b and Qwen2-7b models, when group-wise quantization is applied using INT4 with a group size of 128 (i.e., INT4 g128). PPL indicates the perplexity score on WikiText2, CSR and MMLU refer to the averaged zero-shot accuracies on zero-shot common sense reasoning and MMLU tasks, respectively. * results were quoted from AWQ \cite{awq}. More detailed results are provided in Appendix \ref{Appendix:Detailed Task Results}. 
}
\label{tab:main-4bit}
\end{table}

\section{Experiments}
\subsection{Experimental Setup}
We apply per-group weight quantization under an INT3/INT4, where each group consists of 128 channels. Similar to AWQ\cite{awq}, we use a small calibration set from the Pile dataset\cite{pile} to prevent overfitting to any specific downstream domain. For LLMs, we employ open-source models, including the LLaMA2 and LLaMA3 \cite{llama2, llama3} and Qwen-2 \cite{qwen2} model families.

To evaluate the quantized LLMs, we report perplexity (PPL) on the WikiText-2 dataset\cite{wikitext2} and use standard evaluation metrics for zero-shot Common Sense Reasoning tasks and zero-shot MMLU benchmark\cite{mmlu}.

We compare ROSAQ with the existing quantization methods, such as \textbf{GPTQ},  \cite{gptq}, \textbf{SpinQuant} \cite{spinquant}, and \textbf{AWQ} \cite{awq}. We also report the ``rotation-less'' baseline run, denoted as \textbf{Mixed} which select salient channels according to the activation magnitudes on the original feature space (i.e., ${\mathbf{R}}=  {\mathbf{I}}$ in Eq. (\ref{eq_weight_quantization_mixed_precision_linear_layer})).

\subsection{Main Results}
Table \ref{tab:main-4bit} presents the performances of quantized models in INT4, in terms of the perplexity on WikiText2 and the zero-shot accuracies on Common Sense Reasoning tasks and zero-shot MMLU.

As seen in Table \ref{tab:main-4bit}, ROSAQ exhibits slightly superior to GPTQ\cite{gptq}, SpinQuant\cite{spinquant} and AWQ\cite{awq} except perplexity. Notably in Table \ref{tab:Llama3-total} which is more aggressive 3bit setting, ROSAQ outperforms other methods achieving the highest MMLU results across the various categories. 

Detailed results are presented in Appendix \ref{Appendix:Detailed Task Results}, \ref{Appendix:Kernel Fusion}. In particular, Table \ref{tab:speedup} compares inference throughputs, reporting that ROSAQ achieves approximately 2.3x speedup over FP16 when generating 256 tokens with a batch size of 64.

\section{Conclusion}

In this paper, we proposed ROSAQ, a novel saliency-aware quantization method that uses PCA-based rotation to identify salient channels and preserve them in a high-precision. Experimental results on LLaMA and Qwen-2 models demonstrate that ROSAQ achieves promising outcomes in most cases. In the future, we aim to further generalize rotational invariance by incorporating scaling factors for selecting salient channels.
We also would like to extend saliency-aware quantization to both weights and activations, thereby optimizing the efficiency of the retrieval-augmented generation (RAG) \cite{rag}.

\section*{Limitations}
In ROSAQ, we focus solely on weight quantization, however, weight-activation quantization is often important, especially for retrieval-augmented generation (RAG). Additionally, while we currently use mixed precision by retaining FP16 for salient channels, it would be valuable to explore the use of fully low-precision formats, as in AWQ \cite{awq}, by further generalizing the rotational invariance applied in our approach.

While ROSAQ achieve better performance by explicitly separating salient weights, but hardware inefficiencies still remain. In AWQ \cite{awq}, they fully address the mixed-precision issue by quantizing all weights to 4 bits, making the approach more hardware-friendly. Considering the use of mixed precision, we may need to explore more hardware-friendly approaches to support mixed precision, or generalize the proposed framework by quantizing all weights into low-precision formats.

We explore rotational invariance to select salient features; however, further analysis is needed to understand how this saliency-aware quantization relates to other rotation-based approaches for mitigating outliers. Both theoretical and empirical investigations are required to establish a connection between saliency-aware quantization and outlier-free quantization when rotational operations are adopted.

\section*{Acknowledgments}

\bibliography{custom}

\appendix
\section{ROSAQ vs. AWQ: Rotational invariance vs. Scaling invariance for Selecting Salient Channels}\label{sec:awq_comparision}
ROSAQ and AWQ uses different types of computational invariance of a linear layer -- the rotational and scaling invariance. In other words, while ROSAQ uses the rotational invariance based on Eq. (\ref{eq_rotational_invarience_linear_layer}), 
AWQ \cite{awq} uses the scaling invariance, formulated as follows:
\begin{eqnarray}
{\mathbf{X}}{\mathbf{W}} \approx \left( {\mathbf{X}}{\mathbf{D}}^{-1}\right) Q\left( {\mathbf{D}}{\mathbf{W}} \right) 
\label{scaling_invarience_linear_layer}
\end{eqnarray}
where ${\mathbf{D}} \in {\mathbb{R}}^{d \times d}$ is a positive \textit{diagonal} matrix, where AWQ uses the power of the average magnitudes of channels.

\section{Impact of  Salient Channel Identification}\label{sec:appendix}
\begin{table}[!hbt]
\centering
\begin{tabular}{ll}
\hline
\textbf{Selected}      & \textbf{PPL} $(\downarrow)$  \\ \hline\hline
\textit{Top}         & \textbf{5.57} \\
\textit{Bottom}          & 6.63          \\
\textit{Random}        & 6.59          \\
\textit{Top} \& \textit{Bottom} & 5.58          \\ \hline
\end{tabular}
\caption{Comparison of  variants of the salient channel identification in ROSAQ, in terms of perplexity (PPL) on WikiText-2, under Llama2-7B setting.}
\label{tab:salient_control}
\end{table}

As in Section \ref{section_salient_channel_identification}, after performing PCA, ROSAQ selects the principal dimensions corresponding to the $K$ largest eigenvalues as salient features. To examine the effect of using the top-$K$ eigenvalues, Table \ref{tab:salient_control} compares different methods for identifying salient channels in the PCA-projected space, measured by perplexity (PPL) on WikiText2 under the LLaMA2-7B setting. \textit{Top} refers to ROSAQ’s method of selecting the largest eigenvalues, \textit{Bottom} represents a scheme that selects the least-principal dimensions corresponding to the $K$ smallest eigenvaluesl, \textit{Random} refers to a selection scheme of a random group of $K$-consecutive channels, and \textit{Top} \& \textit{Bottom} refers to a combined scheme that takes both channels selected by \textit{Top} and \textit{Bottom}.
The results show that the \textit{Top} method yields the lowest PPL score, confirming that ROSAQ’s selection scheme, based on the largest eigenvalues, effectively identifies informative salient channels.

\section{Impact of Head-Wise PCA Projection}\label{Appendix:Headwise rotation}
\begin{table}[]
\centering
\begin{tabular}{ll}
\hline
\textbf{Rotation} & \textbf{PPL} $(\downarrow)$  \\ \hline
w/ Head-wise      & 5.57 \\
w/o Head-wise     & 5.94 \\ \hline
\end{tabular}
\caption{Comparison results between the head-wise PCA and the global PCA (corresponds to the row named ``w/o Head-wise row'') in terms of PPL scores on WikiText2 under the LLaMA2-7B setting.}
\label{tab:head-wise}
\end{table}

As in Section \ref{sec:Head-wise}, ROSAQ applies a head-wise PCA projection, where PCA is performed separately on each head-specific attention representation. To examine the effect of head-wise PCA, we explore an alternative approach using global PCA, applied to the concatenated multi-head attention representations, as follows
\begin{eqnarray}
&&\mbox{MHSA}\left({\mathbf{Z}}_{l-1} \right) \approx \nonumber \\
&&\mbox{concat}\left[ {\mathbf{H}}_1, \cdots, {\mathbf{H}}_H \right] {\mathbf{R}}\mbox{ } Q\left( {\mathbf{R}}^T {\mathbf{W}}^O  \right)
\label{eq_global_head_PCA}
\end{eqnarray}
where ${\mathbf{R}} \in {\mathbb{R}}^{d_h \cdot H \times d}$ is the global PCA projection matrix which is obtained by performing the PCA on the calibration set ${\mathbf{X}}_{multi}$ that consists of concatenated multi-head representations, and ${\mathbf{W}}^O \in {\mathbb{R}}^{d_h \cdot H \times d}$ is the global output weight matrix. 

Table \ref{tab:head-wise} presents a comparison between the head-wise PCA and the global PCA (corresponds to the row named ``w/o Head-wise row'') in terms of PPL scores on WikiText2 under the LLaMA2-7B setting. The results show that head-wise PCA achieves a PPL of 5.57, improving upon the PPL of 5.94 from global PCA, confirming the effectiveness of the head-wise PCA approach.

\section{Evaluation of Throughput using Kernel Fusion}\label{Appendix:Kernel Fusion}

To evaluate the throughput of ROSAQ for inference time, we apply the QUICK\cite{QUICK} kernel to assess its advantages over the FP16 implementation in an INT4 quantization setting. Considering that most quantization methods present similar inference speeds to those using FP16 as batch size increases, we also evaluate the throughput using batch sizes ranging from 64 to 128 on an RTX A6000. 

For each test, we design our experiment such that the number of tokens to be generated is the same as the number of the context tokens (e.g., 32 tokens are generated after seeing 32 context tokens generated 32 tokens, and 64 tokens are generated after seeing 64 context tokens). Generation process continues until an Out-of-Memory error is encountered, after which we calculates throughput performance as follows:

\begin{equation}
    \mathrm{DecodeSpeed} = \frac{1}{\mathrm{MedianTimePerToken}},
\end{equation}
\begin{equation}
    \mathrm{Throughput} = \mathrm{DecodeSpeed} \times \mathrm{BatchSize},
\end{equation}

Table \ref{tab:speedup} shows results of throughput for inference, comparing ROSAQ with AWQ\footnote{https://github.com/casper-hansen/AutoAWQ} and GPTQ\footnote{https://github.com/AutoGPTQ/AutoGPTQ}, evaluated on sample sets in Wikitext2, under LLaMA2-7b, with the relative speedup over the FP16 implementation. It is shown that ROSAQ achieves about 2.3x speedup for generating 256 tokens with a batch size of 64, and about 2x speedup for generating 128 tokens with a batch size of 128. 

\begin{table*}[]
\centering
\resizebox{\textwidth}{!}{
\begin{tabular}{ccc|c|cc|cc|cc}
\hline
\multirow{2}{*}{\textbf{Device}} &
  \multirow{2}{*}{\textbf{Batch Size}} &
  \multirow{2}{*}{\textbf{\begin{tabular}[c]{@{}c@{}}\# of given,\\ \# to generate\end{tabular}}} &
  \textbf{FP16} &
  \multicolumn{2}{c|}{\textbf{AWQ}} &
  \multicolumn{2}{c|}{\textbf{GPTQ}} &
  \multicolumn{2}{c}{\textbf{ROSAQ}} \\ \cline{4-10} 
                           &                      &     & Decode Speed & Decode Speed & vs FP16 & Decode Speed & vs FP16 & Decode Speed & \multicolumn{1}{l}{vs FP16} \\ \hline
\multirow{6}{*}{RTX A6000} & \multirow{3}{*}{64}  & 128 & 1106         & 2323         & 110\%   & 1331         & 20\%    & 1667         & 50\%                        \\
                           &                      & 256 & 708          & 1909         & 169\%   & 800          & 12\%    & 1638         & 131\%                       \\
                           &                      & 512 & OOM          & 1395         & -       & 433          & -       & 1235         & -                           \\ \cline{2-10} 
                           & \multirow{3}{*}{128} & 64  & 2077         & 2635         & 26\%    & 2274         & 9\%     & 2788         & 34\%                        \\
                           &                      & 128 & 1299         & 2479         & 90\%    & 1453         & 12\%    & 2604         & 100\%                       \\
                           &                      & 256 & OOM          & 2012         & -       & 806          & -       & 2082         & -                           \\ \hline
\end{tabular}
}
\caption{Comparison of throughput for inference, between ROSAQ, AWQ, and GPTQ, computed on a RTX A6000, using sample sets in Wikitext2, under  LLaMA2-7b, on RTX A6000, with the relative speedup over the FP16 implementation. For kernel funsions, AWQ, GPTQ, and ROSAQ use AWQ kernel, the marlin kernel, and QUICK kernel, respectively.}
\label{tab:speedup}
\end{table*}

\section{Detailed Experiment Results}\label{Appendix:Detailed Task Results}

Tables \ref{tab:Llama2-total}, \ref{tab:Llama3-total}, and \ref{tab:qwen2-total} present the full comparison results of ROSAQ, GPTQ\cite{gptq}, SpinQuant\cite{spinquant}, AWQ\cite{awq}, and Mixed, in terms of PPL on WikiText2, zero-shot accuracies on Common Sense Reasoning and MMLU\cite{mmlu} tasks, under INT4/INT3 quantization, using LLaMA2-7b, LLaMA3-8b, and Qwen2-7b, respectively. Zero-shot commonsense reasoning tasks include PIQA\cite{piqa}, WinoGrande\cite{winogrande}, HellaSwag\cite{hellaswag}, ARC-easy and ARC-challenge \cite{arc}, CommonsenseQA\cite{cqa}, OpenbookQA\cite{obqa}.

\begin{table*}[]
\centering
\resizebox{\textwidth}{!}{
\begin{tabular}{c|cc|c|cccccccc|ccccc}
\hline
\multirow{2}{*}{Model} &
  \multicolumn{2}{c|}{\multirow{2}{*}{Method}} &
  \multirow{2}{*}{PPL $\downarrow$} &
  \multicolumn{8}{c|}{Common Sense Reasoning 0-shot $\uparrow$} &
  \multicolumn{5}{c}{MMLU 0-shot $\uparrow$} \\ \cline{5-17} 
 &
  \multicolumn{2}{c|}{} &
   &
  PIQA &
  WinoGrande &
  HellaSwag &
  ARC\_e &
  ARC\_c &
  CSQA &
  OBQA &
  Avg. &
  Humanities &
  Social Science &
  STEM &
  Other &
  Avg. \\ \hline \hline
\multirow{9}{*}{\begin{tabular}[c]{@{}c@{}}Llama2\\ -7B\end{tabular}} &
  \multicolumn{1}{c|}{FP16} &
  - &
  5.47 &
  79.11 &
  69.3 &
  76 &
  74.58 &
  46.42 &
  32.84 &
  44.2 &
  60.35 &
  38.83 &
  46.12 &
  34.48 &
  47.02 &
  41.26 \\ \cline{2-17} 
 &
  \multicolumn{1}{c|}{\multirow{5}{*}{\begin{tabular}[c]{@{}c@{}}INT4\\ g128\end{tabular}}} &
  GPTQ &
  *5.69 &
  78.62 &
  68.82 &
  75.33 &
  72.56 &
  44.20 &
  28.83 &
  44 &
  58.91 &
  36.96 &
  42.57 &
  31.72 &
  44.67 &
  38.72 \\
 &
  \multicolumn{1}{c|}{} &
  SpinQuant &
  5.52 &
  77.15 &
  67.96 &
  74.39 &
  70.75 &
  42.75 &
  24.73 &
  42.2 &
  57.13 &
  31.20 &
  34.97 &
  28.10 &
  37.46 &
  32.72 \\
 &
  \multicolumn{1}{c|}{} &
  AWQ &
  5.60 &
  79.05 &
  68.67 &
  75.31 &
  73.70 &
  44.28 &
  27.76 &
  44.2 &
  59 &
  36.20 &
  46.70 &
  35.81 &
  46.31 &
  40.64 \\
&
  \multicolumn{1}{c|}{} &
  Mixed &
  5.68 &
  79.22 &
  69.53 &
  75.36 &
  74.16 &
  44.71 &
  31.78 &
  42.6 &
  59.62 &
  37.79 &
  43.74 &
  33.62 &
  45.67 &
  39.90 \\
 &
  \multicolumn{1}{c|}{} &
  Ours &
  5.57 &
  78.62 &
  69.38 &
  75.32 &
  73.19 &
  44.80 &
  31.12 &
  42.6 &
  59.29 &
  39.09 &
  46.47 &
  34.51 &
  46.25 &
  41.26 \\ \cline{2-17} 
 &
  \multicolumn{1}{c|}{\multirow{5}{*}{\begin{tabular}[c]{@{}c@{}}INT3\\ g128\end{tabular}}} &
  GPTQ &
  *6.43 &
  76.66 &
  66.93 &
  70.62 &
  63.97 &
  39.07 &
  22.36 &
  41.2 &
  54.40 &
  28.25 &
  30.03 &
  27.40 &
  30.29 &
  28.90 \\
 &
  \multicolumn{1}{c|}{} &
  SpinQuant &
  6.25 &
  75.90 &
  66.38 &
  70.09 &
  65.24 &
  39.51 &
  21.46 &
  38.8 &
  53.91 &
  26.06 &
  27.66 &
  25.37 &
  30.54 &
  27.25 \\
 &
  \multicolumn{1}{c|}{} &
  AWQ &
  6.24 &
  77.37 &
  68.19 &
  73.32 &
  69.53 &
  44.2 &
  26.37 &
  40 &
  57 &
  31.16 &
  32.27 &
  30.26 &
  37.17 &
  32.53 \\
 &
 \multicolumn{1}{c|}{} &
  Mixed &
  6.30 &
  77.48 &
  68.11 &
  73.20 &
  71.63 &
  42.49 &
  26.78 &
  41 &
  57.24 &
  34.43 &
  39.42 &
  33.42 &
  40.52 &
  36.65 \\
 &
  \multicolumn{1}{c|}{} &
  Ours &
  6.08 &
  78.24 &
  66.46 &
  73.26 &
  70.58 &
  42.92 &
  28.26 &
  39.8 &
  57.07 &
  34.52 &
  40.43 &
  32.76 &
  42.45 &
  37.17 \\ \hline
\end{tabular}%
}
\caption{
Comparison results between ROSAQ and other quantization methods on \textbf{LLaMA2-7B}, when group-wise quantization is applied using INT4/INT3 with a group size of 128 (i.e., INT4/3 g128). PPL indicates the perplexity score on WikiText2, the performances reported at Common Sense Reasoning and MMLU indicate the zero-shot accuracies across tasks. * results were quoted from AWQ \cite{awq}.
}
\label{tab:Llama2-total}
\end{table*}

\begin{table*}[]
\centering
\resizebox{\textwidth}{!}{%
\begin{tabular}{c|cc|c|cccccccc|ccccc}
\hline
\multirow{2}{*}{Model} &
  \multicolumn{2}{c|}{\multirow{2}{*}{Method}} &
  \multirow{2}{*}{PPL ($\downarrow$)} &
  \multicolumn{8}{c|}{Common Sense Reasoning 0-shot ($\uparrow$)} &
  \multicolumn{5}{c}{MMLU 0-shot ($\uparrow$)} \\ \cline{5-17} 
 &
  \multicolumn{2}{c|}{} &
   &
  PIQA &
  WinoGrande &
  HellaSwag &
  ARC\_e &
  ARC\_c &
  CSQA &
  OBQA &
  Avg. &
  Humanities &
  Social Science &
  STEM &
  Other &
  Avg. \\ \hline\hline
\multirow{9}{*}{\begin{tabular}[c]{@{}c@{}}Llama3\\ -8B\end{tabular}} &
  \multicolumn{1}{c|}{FP16} &
  - &
  6.14 &
  80.85 &
  72.61 &
  79.16 &
  77.69 &
  53.33 &
  69.53 &
  45 &
  68.31 &
  54.84 &
  72.96 &
  53.50 &
  71.07 &
  62.10 \\ \cline{2-17} 
 &
  \multicolumn{1}{c|}{\multirow{5}{*}{\begin{tabular}[c]{@{}c@{}}INT4\\ g128\end{tabular}}} &
  GPTQ &
  *6.5 &
  80.03 &
  73.88 &
  77.64 &
  76.81 &
  51.45 &
  65.03 &
  44.4 &
  67.03 &
  53.73 &
  70.39 &
  50.17 &
  68.55 &
  59.86 \\
 &
  \multicolumn{1}{c|}{} &
  SpinQuant &
  6.71 &
  79.22 &
  72.61 &
  78.42 &
  73.7 &
  50.26 &
  66.01 &
  43.2 &
  66.2 &
  52.20 &
  69 &
  50.52 &
  68.04 &
  59 \\
 &
  \multicolumn{1}{c|}{} &
  AWQ &
  6.53 &
  80.52 &
  70.88 &
  77.73 &
  78.11 &
  49.32 &
  80.99 &
  43.6 &
  68.74 &
  53.20 &
  71.47 &
  52.24 &
  68.73 &
  60.48 \\
 &
 \multicolumn{1}{c|}{} &
  Mixed &
  6.63 &
  79.81 &
  72.45 &
  78.19 &
  77.61 &
  52.22 &
  66.09 &
  44.6 &
  67.28 &
  53.43 &
  70.30 &
  50.59 &
  68.39 &
  59.80 \\
 &
  \multicolumn{1}{c|}{} &
  Ours &
  6.50 &
  81.01 &
  70.72 &
  78.21 &
  75.84 &
  51.54 &
  79.36 &
  45.4 &
  68.87 &
  54.24 &
  71.30 &
  52.27 &
  69.75 &
  60.97 \\ \cline{2-17} 
 &
  \multicolumn{1}{c|}{\multirow{5}{*}{\begin{tabular}[c]{@{}c@{}}INT3\\ g128\end{tabular}}} &
  GPTQ &
  *8.2 &
  73.18 &
  67.56 &
  70.95 &
  60.48 &
  39 &
  34.15 &
  37.2 &
  54.65 &
  41.11 &
  49.95 &
  35.65 &
  48.60 &
  43.48 \\
 &
  \multicolumn{1}{c|}{} &
  SpinQuant &
  7.98 &
  78.18 &
  68.98 &
  74.67 &
  73.95 &
  47.78 &
  50.78 &
  40.4 &
  62.11 &
  43.91 &
  49.24 &
  38.03 &
  51.69 &
  45.48 \\
 &
  \multicolumn{1}{c|}{} &
  AWQ &
  8.24 &
  78.13 &
  72.14 &
  73.97 &
  73.4 &
  46.16 &
  59.87 &
  42.8 &
  63.78 &
  46.48 &
  59.70 &
  39.52 &
  57.32 &
  50.21 \\
 &
 \multicolumn{1}{c|}{} &
  Mixed &
  10.67 &
  74.32 &
  68.75 &
  71.52 &
  64.27 &
  41.21 &
  47.50 &
  38.8 &
  58.05 &
  42.64 &
  55.83 &
  40.34 &
  58.42 &
  48.50 \\
 &
  \multicolumn{1}{c|}{} &
  Ours &
  8.11 &
  77.64 &
  68.59 &
  74.1 &
  69.82 &
  45.65 &
  58.64 &
  42 &
  62.35 &
  47.55 &
  60.42 &
  45.83 &
  58.55 &
  52.41 \\ \hline
\end{tabular}%
}
\caption{
Comparison results between ROSAQ and other quantization methods on \textbf{LLaMA3-8B}, when group-wise quantization is applied using INT4/INT3 with a group size of 128 (i.e., INT4/3 g128). PPL indicates the perplexity score on WikiText2, the performances reported at Common Sense Reasoning and MMLU indicate the zero-shot accuracies across tasks. * results were quoted from AWQ \cite{awq}.
}
\label{tab:Llama3-total}
\end{table*}

\begin{table*}[]
\centering
\resizebox{\textwidth}{!}{%
\begin{tabular}{c|cc|c|cccccccc|ccccc}
\hline
\multirow{2}{*}{Model} &
  \multicolumn{2}{c|}{\multirow{2}{*}{Method}} &
  \multirow{2}{*}{PPL ($\downarrow$)} & 
  \multicolumn{8}{c|}{Common Sense Reasoning 0-shot ($\uparrow$)} & 
  \multicolumn{5}{c}{MMLU 0-shot ($\uparrow$)} \\ \cline{5-17} 
 &
  \multicolumn{2}{c|}{} &
   &
  PIQA &
  WinoGrande &
  HellaSwag &
  ARC\_e & 
  ARC\_c & 
  CSQA &
  OBQA &
  Avg. &
  Humanities &
  Social Science &
  STEM &
  Other &
  Avg.  \\ \hline\hline
\multirow{7}{*}{\begin{tabular}[c]{@{}c@{}}Qwen2\\ -7B\end{tabular}} &
  \multicolumn{1}{c|}{FP16} &
  - &
  7.14 &
  81.12 &
  72.30 &
  78.80 &
  74.62 &
  49.74 &
  81.65 &
  44.2 &
  68.92 &
  60.72 &
  80.92 &
  64.95 &
  75.73 &
  69.42 \\ \cline{2-17} 
 &
  \multicolumn{1}{c|}{\multirow{4}{*}{\begin{tabular}[c]{@{}c@{}}INT4\\ g128\end{tabular}}} &
  GPTQ &
  7.3 &
  80.41 &
  70.64 &
  77.84 &
  75.76 &
  51.62 &
  79.28 &
  45.8 &
  68.76 &
  58.55 &
  79.46 &
  61.16 &
  74.12 &
  67.26 \\
 &
  \multicolumn{1}{c|}{} &
  AWQ &
  7.33 &
  80.52 &
  70.88 &
  77.73 &
  78.11 &
  49.32 &
  80.99 &
  43.6 &
  68.74 &
  60.77 &
  80.08 &
  63.53 &
  75.09 &
  68.79 \\
 &
 \multicolumn{1}{c|}{} &
  Mixed &
  7.39 &
  80.47 &
  70.48 &
  78.12 &
  74.07 &
  50.60 &
  80.92 &
  43.4 &
  68.29 &
  58.55 &
  80.18 &
  62.54 &
  74.93 &
  67.81 \\
 &
  \multicolumn{1}{c|}{} &
  Ours &
  7.29 &
  81.01 &
  70.72 &
  78.21 &
  75.84 &
  51.54 &
  79.36 &
  45.4 &
  68.87 &
  60.85 &
  79.95 &
  63.5 &
  75.28 &
  68.82 \\ \cline{2-17} 
 &
  \multicolumn{1}{c|}{\multirow{4}{*}{\begin{tabular}[c]{@{}c@{}}INT3\\ g128\end{tabular}}} &
  GPTQ &
  8.04 &
  77.69 &
  67.48 &
  74.44 &
  71.55 &
  46.5 &
  70.93 &
  43.2 &
  64.54 &
  54.35 &
  72.77 &
  54.17 &
  67.36 &
  61.22 \\
 &
  \multicolumn{1}{c|}{} &
  AWQ &
  8.16 &
  79.71 &
  68.82 &
  74.69 &
  74.96 &
  50.34 &
  77.97 &
  43.2 &
  67.1 &
  55.49 &
  75.92 &
  56.11 &
  71.13 &
  63.57 \\
 &
 \multicolumn{1}{c|}{} &
  Mixed &
  22.25 &
  70.40 &
  58.96 &
  54.77 &
  61.36 &
  42.24 &
  74.94 &
  41.8 &
  57.78 &
  55.47 &
  75.37 &
  57.98 &
  70.10 &
  63.63 \\
 &
  \multicolumn{1}{c|}{} &
  Ours &
  7.97 &
  79.71 &
  69.85 &
  76.58 &
  69.15 &
  46.42 &
  78.13 &
  43.6 &
  66.21 &
  56.37 &
  77.87 &
  61.08 &
  72.22 &
  65.65 \\ \hline
\end{tabular}%
}
\caption{
Comparison results between ROSAQ and other quantization methods on \textbf{Qwen2-7B}, when group-wise quantization is applied using INT4/INT3 with a group size of 128 (i.e., INT4/3 g128). PPL indicates the perplexity score on WikiText2, the performances reported at Common Sense Reasoning and MMLU indicate the zero-shot accuracies across tasks. * results were quoted from AWQ \cite{awq}.
}
\label{tab:qwen2-total}
\end{table*}

\begin{table*}[]
\centering
\resizebox{0.95\textwidth}{!}{
\begin{tabular}{llcccccccccc}
\hline
\multicolumn{2}{c}{\multirow{2}{*}{\textbf{Method}}} & \multicolumn{10}{c}{\textbf{\# Rank of Avg. Magnitude}}                                \\ \cline{3-12} 
\multicolumn{2}{c}{}                        & 1     & 2    & 3    & 4    & 5    & 6    & 7    & 8    & 9    & 10   \\ \hline\hline
\multirow{3}{*}{Rotated}   & Magnitude      & 42.13 & 9.38 & 6.25 & 5.44 & 5.06 & 4.38 & 4.04 & 4.03 & 3.96 & 3.27 \\
                           & Channel Index  & 0     & 1    & 2    & 4    & 3    & 5    & 6    & 8    & 7    & 10   \\ 
                           & Eigenvalue     & 20449.56 & 7449.92 & 4075.54 & 2847.99 & 3839.25 & 2720.07 & 2446.66 & 2038.99 & 2229.96 & 1551.73 \\ \hline
\multirow{2}{*}{Original}  & Magnitude      & 7.44  & 2.12 & 1.23 & 1.21 & 0.88 & 0.70 & 0.65 & 0.62 & 0.51 & 0.50 \\
                           & Channel Index  & 1512  & 2944 & 2298 & 2393 & 3135 & 3431 & 310  & 3893 & 2077 & 1415 \\ \hline
\end{tabular}
}
\caption{Comparison of statistics of the salient channels between the original and PCA-projected spaces, using the LLaMA2-7B model with a random 2048-token sequence from WikiText-2. 1) The average magnitudes of top salient channels, where ten channels are selected after sorting all channels based on their average magnitudes in each respective space. 2) The corresponding eigenvalues in the PCA-projected channels
}
\label{tab:magnitude}
\end{table*}

\section{Activation Values of Salient channels: Original vs. PCA-Projected Spaces}\label{Appendix:Distribution}

To examine how the average magnitudes of the salient channels differ between the original and PCA-projected spaces, Table \ref{tab:magnitude} represents the average magnitudes of top salient channels, where ten channels are selected after sorting all channels based on their average magnitudes in each respective space, using the LLaMA2-7B model with a random 2048-token sequence from WikiText-2. In the PCA-projected channels, we also present their corresponding eigenvalues.

The results show that the PCA-projected space tends to have larger average magnitudes compared to those of the original space. The relative ratio of the average magnitudes between the top-1 and top-2 channels is more dominant in the PCA-projected space compared to the original space.

\section{Weight Splitting Section}\label{Appendix:weightsplitting}
For the attention block, we allocate 128 salient features for $W_Q, W_K$, and $W_V$. In the case of $W_O$, we allocate 32 salient features per head to ensure the group size for quantization is a multiple of 32. For the FFN block, 128 salient features are set for both $W_U$ and $W_G$. However, since the rotation matrix between weights of FFN block cannot be absorbed into $W_U$ and $W_G$, we do not rotate the $W_D$. Instead, we apply per-channel scaling, proposed by AWQ\cite{awq}, to protect the salient features of the $W_D$ during quantization. We also provide an ablation study in Table \ref{tab:salient_control} to test whether the earlier features of the weights are truly salient.

\end{document}